\documentclass[10pt,twocolumn,letterpaper]{article}

\usepackage{lineno} 
\usepackage{iccv}              

\usepackage{multirow}

\usepackage{graphicx}   
\usepackage{booktabs}   
\usepackage{array}      
\usepackage{comment}
\usepackage{capt-of}

\definecolor{iccvblue}{rgb}{0.21,0.49,0.74}
\usepackage[pagebackref,breaklinks,colorlinks,allcolors=iccvblue]{hyperref}

\def\paperID{*****} 
\def\confName{ICCV}
\def\confYear{2025}

\title{VQArt-Bench: A semantically rich VQA Benchmark for Art and Cultural Heritage}

\author{
Andrea Alfarano\thanks{Equal contributors.}\; \thanks{Corresponding author: \texttt{andrea.alfaran@uzh.ch}} \quad
Lorenzo Venturoli\footnotemark[1] \quad
Darío Negueruela del Castillo\\
University of Zurich, Max Planck Society \\
{\tt\small andrea.alfaran@uzh.ch \quad lorenzo.venturoli@uzh.ch \quad dario.neguerueladelcastillo@uzh.ch}
}

\begin{document}


\makeatletter
\let\@oldmaketitle\@maketitle
\renewcommand{\@maketitle}{%
  \@oldmaketitle
  \begin{center}
    \includegraphics[width=\linewidth]%
            {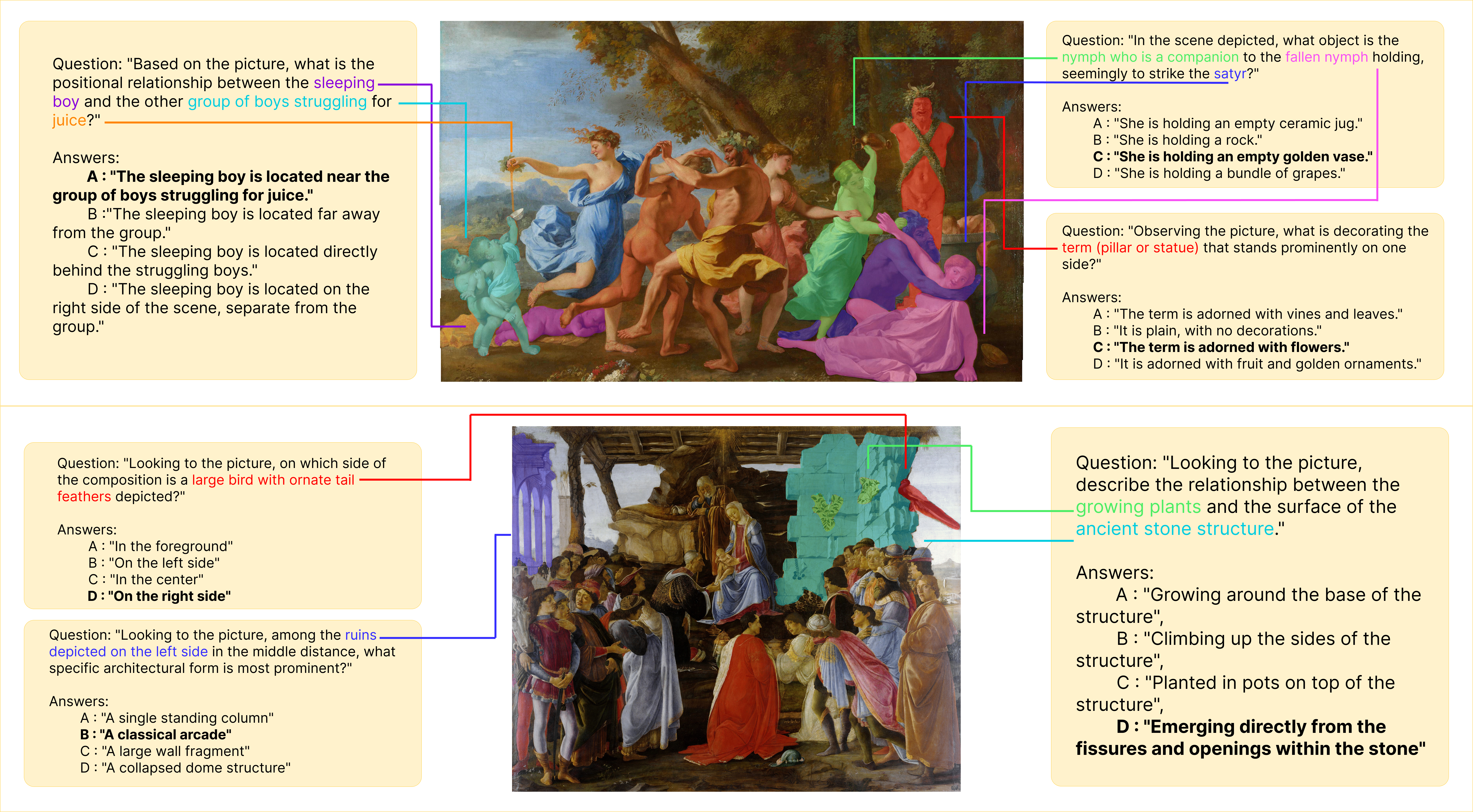}

    \captionof{figure}{Examples from our VQArt-bench with highlighted related subjects for better visualization. VQArt-bench is deeply grounded in the related images
      and refers to specific elements or areas of the artwork. Each
      question requires a profound visual understanding to be
      answered correctly.}
  \end{center}\bigskip
}
\makeatother

\maketitle

\footnotetext[1]{Dataset link: \url{https://github.com/AlfaranoAndrea/VQArt-Bench}.}

\begin{abstract}
Multimodal Large Language Models (MLLMs) have demonstrated significant capabilities in joint visual and linguistic tasks. However, existing Visual Question Answering (VQA) benchmarks often fail to evaluate deep semantic understanding, particularly in complex domains like visual art analysis.  Confined to simple syntactic structures and surface-level attributes, these questions fail to capture the diversity and depth of human visual inquiry. This limitation incentivizes models to exploit statistical shortcuts rather than engage in visual reasoning. 
To address this gap, we introduce VQArt-Bench, a new, large-scale VQA benchmark for the cultural heritage domain. This benchmark is constructed using a novel multi-agent pipeline where specialized agents collaborate to generate nuanced, validated, and linguistically diverse questions. The resulting benchmark is structured along relevant visual understanding dimensions that probe a model's ability to interpret symbolic meaning, narratives, and complex visual relationships.  Our evaluation of 14 state-of-the-art MLLMs on this benchmark reveals significant limitations in current models, including a surprising weakness in simple counting tasks and a clear performance gap between proprietary and open-source models. Our dataset is available here\footnotemark
\end{abstract}

\section{Introduction}
\label{sec:intro}




\begin{figure*}[h!]

        \centering
        \includegraphics[width=\linewidth]{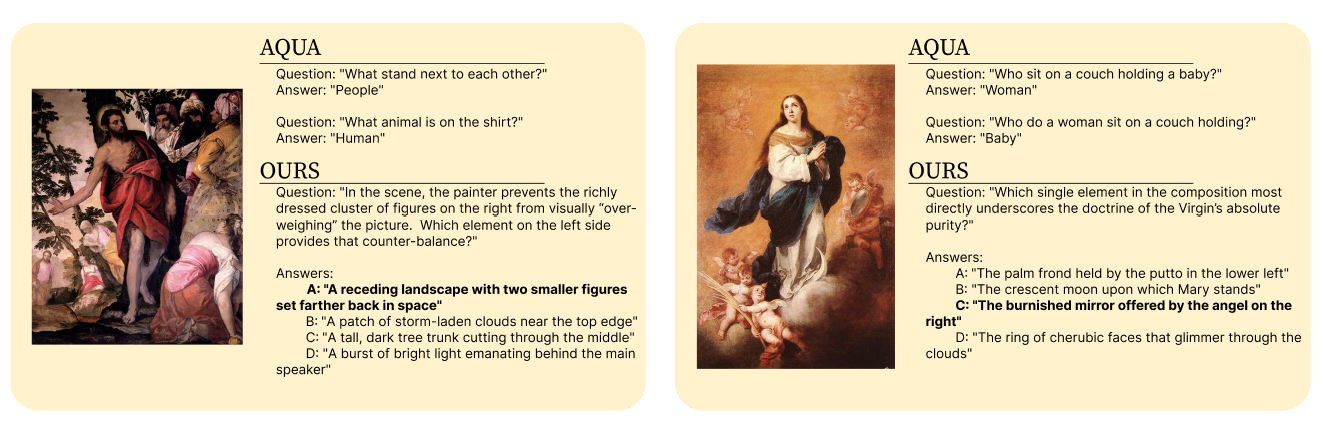}
        \caption{Demonstration of question quality improvement using our agentic pipeline (\textit{Ours}) over a rule-based system (\textit{AQUA}). All the questions have been generated from the same data source. While the rule-based questions are often shallow, lack nuance, and can be factually inconsistent, our method produces context-aware questions appropriate for fine-grained analysis.  The rule-based approach produces semantically shallow questions that lack correct terminology (e.g., referring to the Madonna as a "woman") and can introduce factual inaccuracies, such as hallucinating an "animal on the shirt". In contrast, our agentic pipeline leverages LLMs to generate questions that are both challenging and precise. It preserves the nuances of the source material, formulating sophisticated questions that require a deep understanding of artistic composition (left example) and complex symbolism (right example). Correct answers are reported in bold text.}
        
        \label{fig:comparison}

\end{figure*}

In recent years, MLLMs \cite{hurst2024gpt} have demonstrated exceptional capabilities in image-text comprehension, understanding, reasoning, and generation. VQA \cite{Antol2015, li2023seedbenchbenchmarkingmultimodalllms} has emerged as a critical benchmark for evaluating such models' capabilities. This task requires a model to answer a textual query about an image, a process that necessitates both linguistic and visual understanding.


Advancing these systems toward a true visual interpretation of art requires benchmarks that reflect the complex skills necessary for artistic analysis \cite{garcia2020datasetbaselinesvisualquestion, becattini2023viscounth}. This task is particularly challenging due to the significant domain shift between the images and textual information used to train large models and the unique language of visual art. 
Art, unlike many other visual domains, demands a sophisticated interpretive engagement. The relation between image and meaning in artistic contexts is complex, multilayered, and inherently non-linear. 
Despite significant progress in multimodal vision-language systems, most approaches still rely on relatively reductive assumptions about visual content: being primarily optimized for object detections, image captioning, or scene classification. 

Constructing such representative benchmarks is fraught with challenges, and existing VQA benchmarks often fail, particularly in specialized domains like art and cultural heritage \cite{garcia2020datasetbaselinesvisualquestion, becattini2023viscounth}.
These failures largely stem from the way large-scale VQA datasets for visual art analysis are constructed. Rule-based generation programmatically creates question-answer pairs by taking structured metadata (like image captions or object labels) and inserting them into a fixed set of sentence templates. 
Although this methodology is highly scalable, its reliance on rigid templates has severe limitations, especially in the boundless artistic domain, filled with its unique subjects, complex allegorical relationships, and symbolic actions. First, it leads to a lack of linguistic diversity as the questions are confined to a narrow range of syntactic structures. Second, it produces shallow questions failing to capture context, symbolism, or implicit relationships. Figure \ref{fig:comparison} exemplifies these limitations by comparing questions generated by our proposed method with those from the rule-based benchmark AQUA \cite{garcia2020datasetbaselinesvisualquestion}, which is the first and only publicly available benchmark for art-focused VQA \cite{bleidt2024artquest}. To avoid cherry-picking, we report the first question from the AQUA test and validation sets and generate a question for the same image using the same captions from SemArt \cite{garcia2018read}. As shown, the rule-based approach lacks correct terminology (e.g., referring to the Madonna as a ”woman”) and can even introduce factual inaccuracies, such as hallucinating an ”animal on the shirt” due to flawed metadata parsing.

Rule-based VQA approaches are hindered by their tendency to create questions and answers focused on simplistic, high-frequency patterns \cite{song2024robust, he2020learning}. This problem arises from generating queries automatically from metadata, leading to an overabundance of questions based on overrepresented common patterns like "What is near X?". In turn, the answers become dominated by common terms; for instance, in the AQUA test set, "Human" and "Person" alone account for more than 30\% of all correct replies. This skewed distribution encourages models to exploit linguistic priors instead of performing true visual analysis \cite{he2020learningsceneborrowingrich, abdelkarim2021exploring}. A robust visual art benchmark must move beyond this, testing a model's capacity to recognize the complex actions, rare subjects, and symbolic content—such as historical or cultural motifs—that define an artwork and are systematically ignored by standard datasets.

To address these challenges 
we move beyond the superficial inquiries of rule-based systems and propose a novel agentic pipeline \cite{durante2024agentaisurveyinghorizons} for visual art question generation. Our framework deconstructs the complex task of question creation into a sequential, multi-agent workflow comprising an agentic data cleaning and four question crafting agents: a Topic Selector, a Question Generator, a Question Refiner, and a final Judge. Initially, the Topic Selector analyzes the image caption to identify candidate question categories that can be posed given the available information. The Question Generator then crafts nuanced, open-ended questions based on these candidates. Subsequently, the Question Refiner converts these into a challenging multiple-choice format with plausible distractors, and finally, every question is validated by the Judge, which makes sure that each question is non-trivial, unambiguously answerable from the image, and linguistically correct.

The main contributions of this work are as follows:

\begin{itemize}
    \item We point out current limitations of existing VQA benchmarks for art, demonstrating how rule-based approaches fail to evaluate deep semantic understanding.
    \item The introduction of a novel, multi-agent pipeline for generating questions that are linguistically complex, context-aware, and designed to test for nuanced visual reasoning.
    \item VQArt-bench, a new large-scale VQA dataset featuring semantically rich questions to rigorously benchmark genuine visual literacy in the cultural heritage domain.
    \item An extensive evaluation of 14 state-of-the-art (SOTA) models on our new benchmark across multiple dimensions, showcasing their capabilities in visual art analysis.
\end{itemize}

\section{Related works}
\label{sec:rel_work}

\paragraph{The Evolution of VQA and Multimodal Models}
Visual Question Answering (VQA) was introduced as a benchmark task to measure a machine's ability to reason about visual content in response to natural language queries \cite{pandey2025quest, wu2017visual}. The field has then grown rapidly, leading to the development of numerous datasets, such as COCO-QA \cite{NIPS2015_831c2f88} and VQA \cite{Antol2015}, and a variety of methods focusing on the joint embedding of visual and textual features \cite{hu2018learning, su2018learning}. A parallel line of work has focused on grounding linguistic concepts to visual data, with large-scale datasets like Visual Genome \cite{krishna2016visual} aiming to create dense, fine-grained alignments between images and text. More recently, the landscape has been reshaped by the success of a new class of MLLMs \cite{yin2306survey, hurst2024gpt, team2023gemini, gemmateam2025gemma3technicalreport, aria2024}. These models leverage the powerful reasoning and generation capabilities of LLMs to achieve state-of-the-art performance in multimodal comprehension.

\paragraph{Benchmarking MLLMs }
The rapid development of MLLMs has created an urgent need for benchmarks that can rigorously evaluate their capabilities. Recent benchmarks have been proposed, with a notable trend toward more objective evaluation formats. For instance, MME \cite{fu2023mme}, MMBench \cite{liu2024mmbenchmultimodalmodelallaround}, and SeedBench \cite{li2023seedbenchbenchmarkingmultimodalllms} introduced multiple choice and true/false questions to mitigate the ambiguity and cost associated with human evaluation or LLM-based scoring. However, many existing benchmarks face limitations: some are constrained to a narrow set of visual understanding skills, while others are relatively small in scale, which can lead to unreliable performance metrics \cite{li2023seedbenchbenchmarkingmultimodalllms}.   

\paragraph{VQA in the Specialized Domain of Cultural Heritage}
The translation of VQA into specialized domains such as cultural heritage presents a unique set of challenges \cite{garcia2020datasetbaselinesvisualquestion, becattini2023viscounth, bongini2020visual}. Artworks contain visual information, such as painting techniques, iconographic symbols and historical styles, that is systematically absent from the natural image datasets typically used to train MLLMs. 
Existing art-specific VQA datasets, such as AQUA \cite{garcia2020datasetbaselinesvisualquestion} and VISCOUNTH \cite{becattini2023viscounth}, tried to tackle the presented challenges relying on rule-based or template-based methods, but ultimately falling short due to the inherent limitations of their methods, as shown in Figure \ref{fig:comparison}, explained in Sec. \ref{sec:intro} and later in this section.

While efforts like AQUA and VISCOUNTH have introduced art-specific VQA datasets, their reliance on rule-based or template driven generation results in narrow, often superficial linguistic structures and symbolic blindness which arise from lack of grounding in the interpretive frameworks that have long structure art historical reasoning.
The complexity of artistic images has been systematically addressed in scholarly traditions for more than a century.  Wölfflin's analysis of visual systems, Riegls's notion of Kunstwollen and Warburg's iconographic atlas (Mnemosyne) highlight how form, gesture and motif encode deep cultural memory. Erwin Panofsky's layered theory of interpretation formalizes this complexity in three levels: iconographic recognition of motifs and forms, iconographic identification of subjects, narratives and symbols, and iconologicals analysis of cultural, ideological and psychological models embedded in artworks.
Computational ontology initiatives (ICON, IICONGRAPH) take up this challenge and offer data models and symbolic representations of these Panofskian levels of analysis, which enable semantic enrichment and structured querying. However, these are not designed to test the interpretative capacities of deep learning models in open-ended, dynamic settings.
Our proposed benchmark complements these ontology-driven approaches, shifting the focus from structured representation to active reasoning.

\paragraph{Current limits in Automatic Question Generation Methodologies} \label{par:limits}
Early methodologies for creating large-scale VQA datasets relied on rule-based systems that applied templates to an image's structured semantic annotations, such as scene graphs \cite{Johnson2017, Hudson2019}. While highly scalable and logically grounded in image content, this process resulted in a lack of linguistic diversity and incentivized models to exploit statistical shortcuts rather than engaging in genuine visual reasoning \cite{Goyal2017, Agrawal2018}. To address this limited expressiveness, subsequent research turned to Neural Question Generation (NQG) models, which promised greater linguistic variety \cite{Mostafazadeh2016} but introduced new challenges, including a critical propensity for "hallucinating" factually ungrounded content \cite{Ji2022}. The introduction of LLMs in this pipeline marks the latest shift, demonstrating a remarkable ability to generate more diverse and seemingly faithful questions \cite{Liu2023}. However, these powerful models do not completely overcome the shortcomings of their predecessors and keep introducing subtle factual inconsistencies or hallucinations that are harder to detect due to their high fluency \cite{Yin2023}; furthermore, their monolithic nature makes the generation process difficult to control or verify.

\begin{figure*}[h!]

        \centering
        \includegraphics[width=\linewidth]{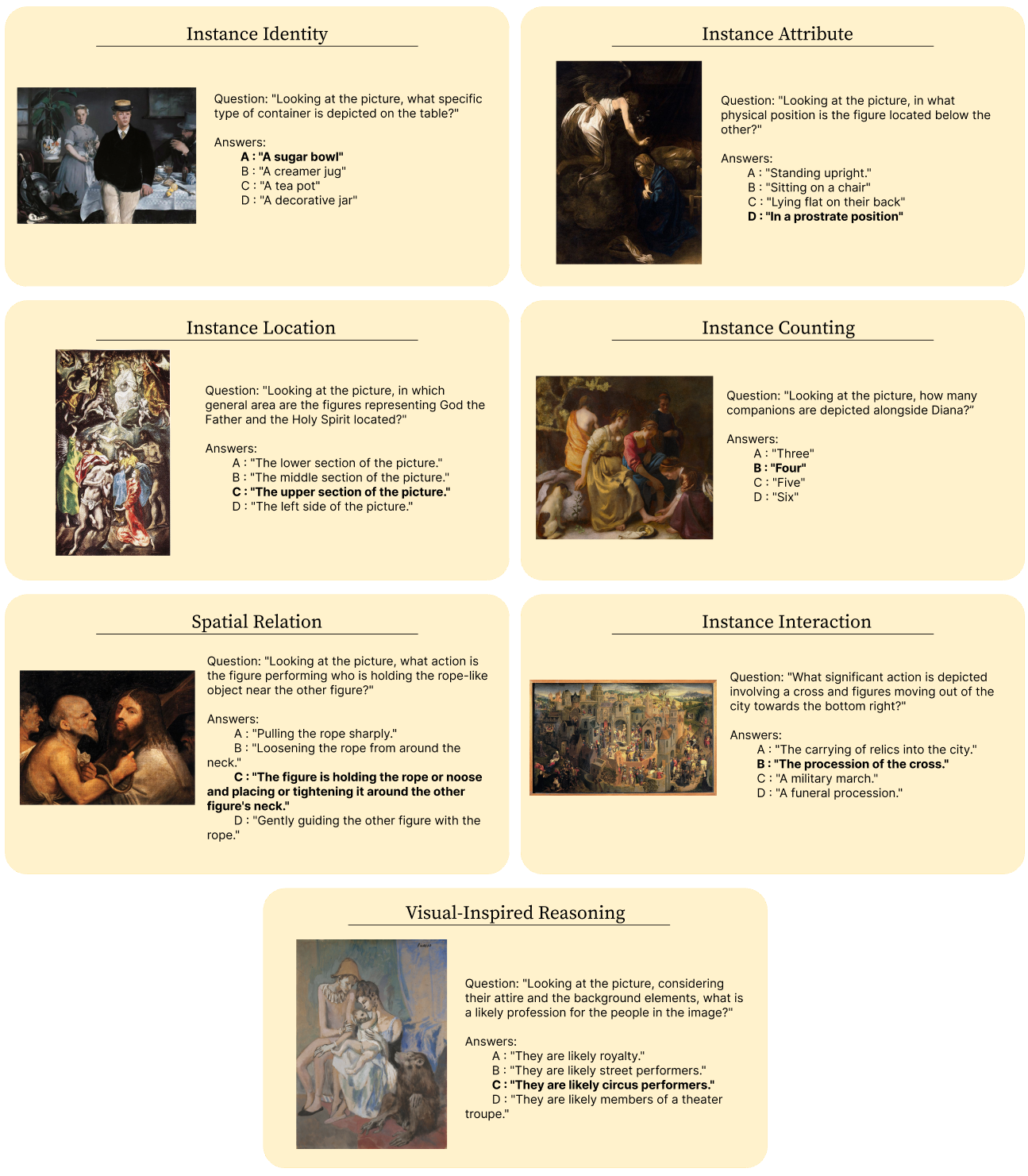}
        \caption{Examples from our \textit{VQArt-Bench} dataset, categorized by our seven core evaluation dimensions. Our benchmark is designed to test a spectrum of visual reasoning skills. It challenges models on foundational abilities like identifying objects and their properties (\textit{Instance Identity}, \textit{Instance Attribute}), locating them in the scene (\textit{Instance Location}), and quantifying them (\textit{Instance Counting}). The evaluation then progresses to more complex compositional tasks, such as understanding \textit{Spatial Relation} and \textit{Instance Interaction}, and high-level tasks that require inferring context and causality (\textit{Visual-Inspired Reasoning}). Correct answers are highlighted in Bold text.}
        \label{fig:categories}
    
\end{figure*}

\begin{figure*}[h!]

        \centering
        \includegraphics[width=0.7\linewidth]{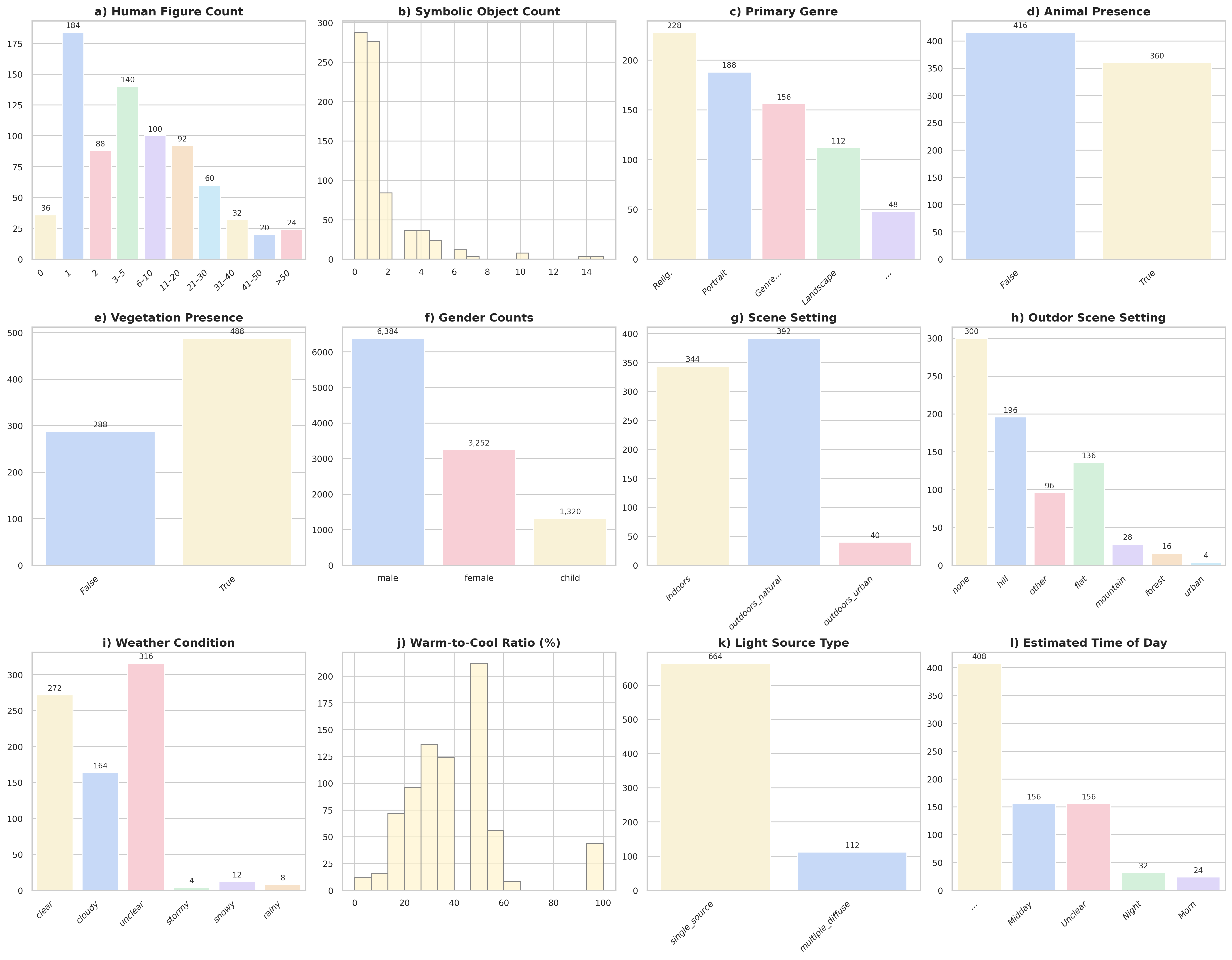}
        \caption{Exploitative LLM based statistical evaluation of key attributes in our \textit{VQArt-Bench}. The dataset shows broad diversity in terms of \textbf{compositional elements}, including (a) a wide range of human figure counts, from individuals to large crowds; (b) varied numbers of symbolic objects per scene; and (f) representation across different genders. The collection spans multiple \textbf{genres and settings}, covering (c) primary artistic genres like religious and portraiture; (d, e) a balance of scenes with and without animal or vegetation presence; and (g, h) a mix of indoor and diverse outdoor environments. Finally, the dataset captures rich \textbf{atmospheric and stylistic variations}, including (l) different times of day; (i) various weather conditions; (k) simple and complex lighting sources; and (j) a full spectrum of warm-to-cool color ratios, indicating diverse visual moods.}
        \label{fig:statistics}
    
\end{figure*}

\section{Methods}

\subsection{Evaluation dimensions}\label{sect_topics}
To ensure that our benchmark evaluates a comprehensive range of visual understanding skills, we use a well-established hierarchy of seven distinct reasoning dimensions derived from \cite{li2023seedbenchbenchmarkingmultimodalllms}. These dimensions, inspired by cognitive science, structure the evaluation from basic perception to complex inference, allowing for a granular analysis of a model's capabilities.

\begin{itemize}
    \item \textbf{Instance Identity:} Involves identifying a specific instance, including its existence or class, based on visual evidence.
    \item \textbf{Instance Attribute:} Relates to the specific visual attributes of an instance, such as its color, shape, texture, or material.
    \item \textbf{Instance Location:} Concerns the absolute or relative position of a specified instance within the image frame.
    \item \textbf{Instance Counting:} Requires the model to accurately count all occurrences of a specific object class.
    \item \textbf{Spatial Relation:} Tests the ability to recognize the relative spatial relationship between two or more distinct objects.
    \item \textbf{Instance Interaction:} Requires recognizing actions or relationships between two or more subjects or objects.
    \item \textbf{Visual-Inspired Reasoning:} Evaluates a model's ability to perform common-sense or causal reasoning based on the visual scene (e.g., inferring intent, cause, or future outcomes).
\end{itemize}
It is possible to see an example for each category in Fig. \ref{fig:categories}.

\subsection{Data Curation and Pre-processing}\label{sect_data_cleaning}
Popular VQA benchmark curation methods propose to directly generate VQA from pictures by tasking an LLM to generate a relevant question given the picture \cite{li2023seedbenchbenchmarkingmultimodalllms}. We found that in visual art analysis, this can lead to superficial questions which assess basic visual appearance due to the difficulty of fully operationalizing what is \textit{relevant}.
An established option is to ground VQA questions on expert-curated descriptions and metadata often present in art repositories, so that the questions cover the same topics and relevant aspects that the curator decided to discuss.
A key challenge in using such data is that it often does not contain only the visual information of the target artwork; rather, it usually interleaves the desired visual analysis with non-visual contextual information, such as artist biographies or historical background. Additionally, the contextual information for one artwork is often based on the descriptions of other artworks (such as preceding artworks by the same artists). This leads to the necessity of isolating visually relevant text, while also distinguishing which visual elements refer to the target work and which refer to external artworks. Such elements, if not filtered out, may lead to hallucinated questions in the following steps. For this purpose, we implemented a pre-processing step using an LLM that, unlike a rigid, rule-based filter that might be erroneously confused by text ambiguity, is able to leverage contextual understanding to parse the raw article and extract only the most relevant sentences that refer to the target artwork. This step yields a clean, relevant textual signal for our generation pipeline.

Our question generation pipeline is designed to be source-agnostic. For this work, we selected Wikipedia as our primary data source due to its vast repository of images accompanied by human-generated collaborative descriptions \cite{Srinivasan2021}; however, other resources can be used effectively. The initial phase involved downloading approximately 30,000 images and their corresponding descriptions. To ensure each description was sufficiently rich for generating detailed questions, we applied a length filter, discarding any image-text pair where the article contained fewer than 400 words. Each image text pair has been cleaned by an LLM. We found that while passing both the picture and the text may improve the ability of the LLM to discriminate which content is actually in the picture, this also can potentially lead to hallucinate descriptions not contained in the text, and that the contextual understanding of the LLM is still sufficient for the task, leaving hallucination checks to next steps in the VQA generation pipeline.

\subsection{Agent-Based Question Generation}
From the cleaned descriptions of \ref{sect_data_cleaning} we generate questions using a sequential pipeline of four specialist agents. This division-of-labor approach ensures questions grounded in the metadata, reduce hallucinations and relevance regarding the desired evaluations dimension of Sec. \ref{sect_topics}

\paragraph{Step 1: Topic Selection and Grounding}
The pipeline begins with the \textbf{Topic Selector} Agent, which is tasked to analyze the pre-processed visual description of each artwork, and propose a list of candidate topics that may be relevant for that picture.  We found that this step is really prone to hallucinate facts to support the relevant questions. 
To better ground each question, for each candidate topic the agents output must cite the minimal text snippet from the cleaned description containing the supporting answer, ensuring that all subsequent generation is based on factual information.

\paragraph{Step 2: Open-Ended Question Formulation}
The grounded topics are then passed to the \textbf{Question Generator}. This agent takes the proposal from Step 1 and formulates nuanced, open-ended questions. While the topic agents primary goals is to ground the question and reduce hallucinations, the open ended agent goal is to formulate the question in the most relevant way, leaving the formulation of the closed ended option to the next agents. The agent is tasked to keep the question \textit{informed} by the text, and answerable \textit{by observing the image}  \cite{romero2024cvqaculturallydiversemultilingualvisual}, avoiding questions that would be answerable only by looking to the metadata.

\paragraph{Step 3: Multiple-Choice Refinement and Distractor Generation}
The open-ended questions are then given to the \textbf{Question Refiner}, which converts them into a challenging multiple-choice format. This agent is explicitly tasked to design highly plausible distractors by anticipating possible visual misinterpretations grounded on the specific image, incorporating subtle details, and using contextually relevant but incorrect information \cite{Gao2019_distractor}. This step is crucial for creating questions that demand deep and precise visual analysis.

\paragraph{Step 4: Final Judgment and Validation}
Finally, every multiple-choice question is evaluated by the \textbf{Judge}. This agent acts as a quality gatekeeper, ensuring that each question is unambiguously answerable from the image, is non-trivial, follows the provided evaluation dimensions and is linguistically sound \cite{Zheng2023_llmjudge}.  Only questions that pass this stringent assessment are included in the benchmark.

\subsection{Human Validation and Quality Assurance}

Although our automated pipeline is designed for high fidelity, we performed a human validation study to rigorously quantify the final quality of our benchmark. We randomly sampled 25\% of the generated image-question pairs for manual review. Expert annotators were tasked with verifying that each question was factually grounded in the corresponding image. This process confirmed that over the 98\% of the reviewed questions were free from hallucinations, underscoring the reliability of our agentic approach.
To make sure that our questions capture the essence of the source descriptions, we compare the generated questions against the original text. The analysis found that in most of the of cases, the questions successfully represented most of the salient information. This result demonstrates that our pipeline not only produces linguistically diverse questions but also achieves high information coverage, rivaling the thoroughness of more rigid methods.

\subsection{Final Benchmark Composition}
The complete \textbf{VQArt-Bench} benchmark consists of 14,463 high-quality, multiple-choice questions that span the seven reasoning dimensions presented above. The final distribution of questions across these categories is detailed in Table \ref{tab:dataset_stats}. This balanced composition ensures that the models are evaluated across a comprehensive spectrum of visual skills.

To provide an approximate quantitative overview of key attributes within VQArt-Bench, we automatically analyzed the dataset's statistical distribution of compositional elements using Gemini 2.5 \cite{team2023gemini} as State-Of-The-Art visual LLM \cite{Liu2023}. This analysis reveals the dataset's broad diversity across several categories and can be seen in Figure \ref{fig:statistics}. The collection shows significant variation in its compositional elements, including: (a) a wide range of human figure counts, with a focus on scenes depicting a small number of individuals (1-3 figures) but also including large crowds; (b) a varied representation across men, women, and children, with a majority presence of male figures \\
The dataset also spans multiple genres and settings, covering: (c) a distribution across primary artistic genres, with Religious art, Portraiture, being the most prominent categories; (d, e) a significant presence of vegetation, while most scenes do not feature animals, providing distinct contexts for analysis; and (g, h) a mix of indoor and outdoor environments, with indoor settings and natural landscapes being more common than urban scenes, and most outdoor scenes set during the day. \\
Finally, the dataset captures rich atmospheric and stylistic variations, including: (l) a variety of estimated times of day, with a focus on midday and afternoon; (i) a range of weather conditions, predominantly featuring clear or cloudy skies; (k) a strong emphasis on artworks with a single, clear light source over more complex, diffuse lighting; and (j) a full spectrum of warm-to-cool color ratios, with a tendency towards balanced or cooler color palettes, indicating diverse visual moods.



\begin{table}[h]
  \centering
  \caption{Distribution of Questions  in the VQArt\textendash Bench benchmark.}
  \label{tab:dataset_stats}

  \resizebox{0.8\linewidth}{!}{%
    \begin{tabular}{lc}
      \hline
      \textbf{Reasoning Dimension} & \textbf{Number of Questions} \\ \hline
      Instance Identity          & 2031  \\
      Instance Attribute         & 2598  \\
      Instance Location          & 2100  \\
      Instance Counting          & 1710  \\
      Spatial Relation           & 2067  \\
      Instance Interaction       & 1794  \\
      Visual-Inspired Reasoning  & 2163  \\ \hline
      \textbf{Total}             & \textbf{14463} \\ \hline
    \end{tabular}%
  }
\end{table}

\begin{table*}[h!] 
  \centering 
  \caption{Evaluation results of different models on VQArt-Bench. Seven spatial–reasoning dimensions and overall accuracy.}%
  \label{tab:verall} 

  \resizebox{\textwidth}{!}{%
  \begin{tabular}{llrrrrrrrr} 
  	\toprule 
  	\multirow{2}{*}{\textbf{Source}} & 
  	\multirow{2}{*}{\textbf{Model}} & 
  	\multicolumn{8}{c}{\textbf{Evaluation Dimensions}} \\ 
  	\cmidrule(lr){3-10} 
  	 &  & 
  	  \textbf{Instance Attribute} & 
  	  \textbf{Instance Localization} & 
  	  \textbf{Instance Counting} & 
  	  \textbf{Spatial Relation} & 
  	  \textbf{Instance Interaction} & 
  	  \textbf{Instance Identity} & 
  	  \textbf{Visual-Inspired Reasoning} & 
  	  \textbf{Overall Acc.} \\ 
  	\midrule 
  	\multirow{5}{*}{Closed Source} 
  	  & Gemini~2.5      & 0.73  & 0.73  & 0.66 & 0.72  & 0.75    & 0.74    & 0.80    & 0.71    \\ 
  	  & GPT-4o      & 0.66  & 0.66  & 0.59 & 0.65  & 0.68    & 0.67    & 0.72    & 0.64    \\ 
  	  & GPT-4o~mini & 0.58  & 0.58  & 0.53 & 0.58  & 0.60    & 0.59    & 0.64    & 0.57    \\ 
  	\midrule 
  	\multirow{6}{*}{Open Source} 
  	  & Aria~\cite{aria2024}                
      & 0.60 & 0.59 & 0.54 & 0.51 & 0.61 & 0.60 & 0.77 & 0.58 \\ 
  	  & Aya~Vision~\cite{ayavision2025}       
      & 0.59 & 0.53 & 0.53 & 0.56   & 0.58   & 0.61   & 0.75   & 0.57   \\ 
  	  & Kimi-VL~\cite{kimivl2025}             
      & 0.69 & 0.70 & 0.64 & 0.66   & 0.67   & 0.72 & 0.83   & 0.67   \\ 
  	  & Phi4~\cite{phi42024}      
      & 0.59 & 0.52 & 0.48 & 0.56 & 0.55 & 0.55 & 0.72 & 0.54 \\ 
  	  & Pixtral~\cite{pixtral2024}                    
      & 0.65 & 0.63 & 0.57 & 0.62 & 0.61 & 0.63 & 0.81 & 0.62 \\ 
  	  & InstructBLIP-Vicuna~\cite{instructblip2023, vicuna2023}       
      & 0.25 & 0.21 & 0.10 & 0.21 & 0.24 & 0.47 & 0.26 & 0.24 \\ 
  	  & LLaVA~\cite{liu2023llava}                      
      & 0.44 & 0.36 & 0.26 & 0.37 & 0.37 & 0.53 & 0.50 & 0.39 \\ 
  	  & LLaVA-NeXT~\cite{liu2024llavanext}                
      & 0.51 & 0.48 & 0.31 & 0.45 & 0.47 & 0.52 & 0.66 & 0.46 \\ 
  	  & Gemma~3~27B~\cite{gemmateam2025gemma3technicalreport}             
      & 0.41 & 0.39 & 0.39 & 0.41 & 0.47 & 0.54 & 0.54 & 0.42 \\ 
  	  & Gemma~3~12B~\cite{gemmateam2025gemma3technicalreport}             
      & 0.38 & 0.33 & 0.33 & 0.33 & 0.38 & 0.42 & 0.47 & 0.36 \\ 
  	  & Gemma~3~4B~\cite{gemmateam2025gemma3technicalreport}              
      & 0.35 & 0.27 & 0.26 & 0.30 & 0.39 & 0.39 & 0.47 & 0.33 \\ 
  	\bottomrule 
  \end{tabular}%
  }
 \end{table*}

\section{Evaluation Results}

\paragraph{Evaluated models} We evaluated 14 models, including 3 variants of Gemma3 \cite{gemmateam2025gemma3technicalreport} in order to test how performances change at scale. We evaluate other open source models as Aria \cite{aria2024}, Aya Vision \cite{ayavision2025}, Kimi-VL \cite{kimivl2025}, Phi-4 \cite{phi42024}, Pixtral 12B \cite{pixtral2024}, LLaVA \cite{liu2023llava}, LLaVA-NeXT \cite{liu2024llavanext}, InstructBLIP-Vicuna \cite{instructblip2023, vicuna2023}, as well as some SOTA closed source models as Gemini 2.5 \cite{team2023gemini}, GPT-4o and GPT-4o mini \cite{hurst2024gpt}.

\subsection{Analysis}
The evaluation results of different models on VQArt-Bench are listed in Table \ref{tab:verall}, where accuracy refers to the proportion of correctly answered multiple-choice questions relative to the total number of questions.
We have observed a number of findings that can inform and guide future work.

\paragraph{Most MLLMs still exhibit limited performance}
As depicted in Tab. \ref{tab:verall}, most of the models analyzed fail to achieve particularly high overall scores, displaying limited capabilities in art understanding. Although most of the models achieve better results than random guessing (four choices: $\sim$25\%), most of them struggle to reach 50\% accuracy.

\paragraph{MLLMs struggle with enumerating instances, while overperform in reasoning}
Table \ref{tab:verall} shows that all evaluated models perform significantly below their overall accuracy in Instance Counting, yet excel in Visual-Inspired Reasoning. This outcome is 
counterintuitive, as a human would find counting elements much simpler than reasoning about an entire artwork. We can explain this phenomenon by looking at the skills required: Visual-Inspired Reasoning demands less specific knowledge about individual instances and a greater capacity for generalization, which is a key strength of current state-of-the-art MLLMs. Furthermore, Instance Counting questions are constructed with far more likely distractors, which easily trick "more naive" models, as it happens with InstructBLIP-Vicuna \cite{instructblip2023, vicuna2023}.

\paragraph{Artistic visual analysis is more challenging then standard benchmarks}
While standard image benchmarks also feature naturalistic landscapes and human interactions, artistic scenes present a unique challenge. They often depict a broader range of subjects beyond everyday life, including historical, religious and fictional figures. The actions portrayed can be far more complex and uncommon than those typically found in online images. Furthermore, the visual appearance of art is fundamentally different due to diverse artistic styles and historical contexts, a stark contrast to the more uniform distribution of images found on the internet, from which large-scale benchmarks are typically sourced. \\
This effect can be seen by looking at table \ref{tab:verall}, more specifically at the results achieved by InstructBLIP-Vicuna, which obtained a score just below the "random guessing" value of 25\%, while achieving the best overall score in the SEEDBench dataset \cite{li2023seedbenchbenchmarkingmultimodalllms}: 59\% with similar evaluation.

\paragraph{Closed source models lead the performances} Our evaluation demonstrates a clear performance disparity between closed-source and open-source models in the domain of artistic visual analysis. Notably, Gemini 2.5 surpasses all current baselines on every evaluation metric. Although open-source models like Kimi-VL are showing promising capabilities, they have not yet reached the same level of performance. This finding underscores the ongoing need for the open-source research community to intensify efforts in developing more capable models tailored to the unique challenges of artistic interpretation.

\paragraph{Kimi-VL is the best Open Source Model} 
From the results in table \ref{tab:verall} it appears that Kimi-VL \cite{kimivl2025} is the best model among all the open source MLLMs, even outperforming closed source models like GPT-4o and GPT-4o mini. This may be due to its native-resolution vision encoder (MoonViT) and its novel training strategy, which directly targets visual reasoning instead of more standard tasks like image captioning. Kimi-VL performances suggest that MoE models \cite{cai2024survey} can be effective in improving performance while reducing active parameters.

\paragraph{Larger models perform generally better:} 
We experimented with how the amount of parameters influences the performance by testing our benchmark against three different versions of Gemma 3 (4B, 12B, 27B) \cite{gemmateam2025gemma3technicalreport}. We observed that, as expected, the same model achieved better results at larger scales, without being subject to fine-tuning of any kind: Gemma 3 - 27B improves by +9\% compared to the 4B version and by +6\% compared to the 12B version. Another example of such is the improved score of GPT-4o when compared to its \textit{mini} version.

\section{Conclusion}
In this work, we addressed the critical limitations of rule-based VQA for art. Such methods, constrained by rigid templates, produce linguistically and semantically shallow questions, leading to skewed data distributions that prevent genuine visual evaluation. To solve this, we introduced VQArt-bench, a new benchmark built with a novel agentic pipeline to generate semantically rich and challenging questions. Our evaluation of 14 state-of-the-art models revealed significant performance limitations, highlighting a surprising weakness in instance counting alongside a strength in visual reasoning. While closed models like Gemini 2.5 currently lead, the promising results from open-source models such as Kimi-VL show a path forward. VQArt-bench effectively exposes current model weaknesses and sets a more rigorous standard for developing AI with genuine visual literacy for art VQA.
{
    \small
    \bibliographystyle{ieeenat_fullname}
    \bibliography{main}

\begin{thebibliography}{46}
\providecommand{\natexlab}[1]{#1}
\providecommand{\url}[1]{\texttt{#1}}
\expandafter\ifx\csname urlstyle\endcsname\relax
  \providecommand{\doi}[1]{doi: #1}\else
  \providecommand{\doi}{doi: \begingroup \urlstyle{rm}\Url}\fi

\bibitem[Abdelkarim et~al.(2021)Abdelkarim, Agarwal, Achlioptas, Chen, Huang, Li, Church, and Elhoseiny]{abdelkarim2021exploring}
Sherif Abdelkarim, Aniket Agarwal, Panos Achlioptas, Jun Chen, Jiaji Huang, Boyang Li, Kenneth Church, and Mohamed Elhoseiny.
\newblock Exploring long tail visual relationship recognition with large vocabulary.
\newblock In \emph{Proceedings of the IEEE/CVF International Conference on Computer Vision}, pages 15921--15930, 2021.

\bibitem[Abdin et~al.(2024)]{phi42024}
Marah Abdin et~al.
\newblock Phi-4 technical report.
\newblock \emph{arXiv preprint arXiv:2412.08905}, 2024.

\bibitem[Agrawal et~al.(2018)Agrawal, Batra, Parikh, and Kembhavi]{Agrawal2018}
Aishwarya Agrawal, Dhruv Batra, Devi Parikh, and Aniruddha Kembhavi.
\newblock Don't just assume; look and answer: Overcoming priors for visual question answering.
\newblock In \emph{Proceedings of the IEEE Conference on Computer Vision and Pattern Recognition (CVPR)}, 2018.

\bibitem[Agrawal et~al.(2024)]{pixtral2024}
Pravesh Agrawal et~al.
\newblock Pixtral 12b.
\newblock \emph{arXiv preprint arXiv:2410.07073}, 2024.

\bibitem[Antol et~al.(2015)Antol, Agrawal, Lu, Mitchell, Batra, Zitnick, and Parikh]{Antol2015}
Stanislaw Antol, Aishwarya Agrawal, Jiasen Lu, Margaret Mitchell, Dhruv Batra, C.~Lawrence Zitnick, and Devi Parikh.
\newblock {VQA}: Visual question answering.
\newblock In \emph{Proceedings of the IEEE International Conference on Computer Vision (ICCV)}, 2015.

\bibitem[Becattini et~al.(2023)Becattini, Bongini, Bulla, Bimbo, Marinucci, Mongiov{\`\i}, and Presutti]{becattini2023viscounth}
Federico Becattini, Pietro Bongini, Luana Bulla, Alberto~Del Bimbo, Ludovica Marinucci, Misael Mongiov{\`\i}, and Valentina Presutti.
\newblock Viscounth: a large-scale multilingual visual question answering dataset for cultural heritage.
\newblock \emph{ACM Transactions on Multimedia Computing, Communications and Applications}, 19\penalty0 (6):\penalty0 1--20, 2023.

\bibitem[Bleidt et~al.(2024)Bleidt, Eslami, and de~Melo]{bleidt2024artquest}
Tibor Bleidt, Sedigheh Eslami, and Gerard de Melo.
\newblock Artquest: Countering hidden language biases in artvqa.
\newblock In \emph{Proceedings of the IEEE/CVF Winter Conference on Applications of Computer Vision}, pages 7326--7335, 2024.

\bibitem[Bongini et~al.(2020)Bongini, Becattini, Bagdanov, and Del~Bimbo]{bongini2020visual}
Pietro Bongini, Federico Becattini, Andrew~D Bagdanov, and Alberto Del~Bimbo.
\newblock Visual question answering for cultural heritage.
\newblock In \emph{IOP Conference Series: Materials Science and Engineering}, page 012074. IOP Publishing, 2020.

\bibitem[Cai et~al.(2024)Cai, Jiang, Wang, Tang, Kim, and Huang]{cai2024survey}
Weilin Cai, Juyong Jiang, Fan Wang, Jing Tang, Sunghun Kim, and Jiayi Huang.
\newblock A survey on mixture of experts.
\newblock \emph{arXiv preprint arXiv:2407.06204}, 2024.

\bibitem[Dai et~al.(2023)Dai, Li, et~al.]{instructblip2023}
Wenliang Dai, Junnan Li, et~al.
\newblock Instructblip: Towards general-purpose vision-language models with instruction tuning.
\newblock \emph{arXiv preprint arXiv:2305.06500}, 2023.

\bibitem[Dash et~al.(2025)]{ayavision2025}
Saurabh Dash et~al.
\newblock Aya vision: Advancing the frontier of multilingual multimodality.
\newblock \emph{arXiv preprint arXiv:2505.08751}, 2025.

\bibitem[Durante et~al.(2024)Durante, Huang, Wake, Gong, Park, Sarkar, Taori, Noda, Terzopoulos, Choi, Ikeuchi, Vo, Fei-Fei, and Gao]{durante2024agentaisurveyinghorizons}
Zane Durante, Qiuyuan Huang, Naoki Wake, Ran Gong, Jae~Sung Park, Bidipta Sarkar, Rohan Taori, Yusuke Noda, Demetri Terzopoulos, Yejin Choi, Katsushi Ikeuchi, Hoi Vo, Li Fei-Fei, and Jianfeng Gao.
\newblock Agent ai: Surveying the horizons of multimodal interaction, 2024.

\bibitem[Fu et~al.(2023)Fu, Chen, Shen, Qin, Zhang, Lin, Qiu, Lin, Yang, Zheng, et~al.]{fu2023mme}
Chaoyou Fu, Peixian Chen, Yunhang Shen, Yulei Qin, Mengdan Zhang, Xu Lin, Zhenyu Qiu, Wei Lin, Jinrui Yang, Xiawu Zheng, et~al.
\newblock Mme: a comprehensive evaluation benchmark for multimodal large language models. corr abs/2306.13394 (2023), 2023.

\bibitem[Gao et~al.(2019)Gao, Tan, Chen, and Kao]{Gao2019_distractor}
Chen-Kai Gao, Chieh-Hsin Tan, Ke-Jia Chen, and Hung-Yu Kao.
\newblock Generating plausible distractors for reading comprehension questions with generation-based and retrieval-based methods.
\newblock In \emph{Proceedings of the 3rd Workshop on Neural Generation and Translation}, 2019.

\bibitem[Garcia and Vogiatzis(2018)]{garcia2018read}
Noa Garcia and George Vogiatzis.
\newblock How to read paintings: semantic art understanding with multi-modal retrieval.
\newblock In \emph{Proceedings of the European Conference on Computer Vision (ECCV) Workshops}, pages 0--0, 2018.

\bibitem[Garcia et~al.(2020)Garcia, Ye, Liu, Hu, Otani, Chu, Nakashima, and Mitamura]{garcia2020datasetbaselinesvisualquestion}
Noa Garcia, Chentao Ye, Zihua Liu, Qingtao Hu, Mayu Otani, Chenhui Chu, Yuta Nakashima, and Teruko Mitamura.
\newblock A dataset and baselines for visual question answering on art, 2020.

\bibitem[Goyal et~al.(2017)Goyal, Khot, Summers-Stay, Batra, and Parikh]{Goyal2017}
Yash Goyal, Tejas Khot, Douglas Summers-Stay, Dhruv Batra, and Devi Parikh.
\newblock Making the {V} in {VQA} matter: Elevating the role of image understanding in visual question answering.
\newblock In \emph{Proceedings of the IEEE Conference on Computer Vision and Pattern Recognition (CVPR)}, 2017.

\bibitem[He et~al.(2020{\natexlab{a}})He, Gao, Song, Cai, and Li]{he2020learning}
Tao He, Lianli Gao, Jingkuan Song, Jianfei Cai, and Yuan-Fang Li.
\newblock Learning from the scene and borrowing from the rich: Tackling the long tail in scene graph generation.
\newblock \emph{arXiv preprint arXiv:2006.07585}, 2020{\natexlab{a}}.

\bibitem[He et~al.(2020{\natexlab{b}})He, Gao, Song, Cai, and Li]{he2020learningsceneborrowingrich}
Tao He, Lianli Gao, Jingkuan Song, Jianfei Cai, and Yuan-Fang Li.
\newblock Learning from the scene and borrowing from the rich: Tackling the long tail in scene graph generation, 2020{\natexlab{b}}.

\bibitem[Hu et~al.(2018)Hu, Chao, and Sha]{hu2018learning}
Hexiang Hu, Wei-Lun Chao, and Fei Sha.
\newblock Learning answer embeddings for visual question answering.
\newblock In \emph{Proceedings of the IEEE Conference on Computer Vision and Pattern Recognition}, pages 5428--5436, 2018.

\bibitem[Hudson and Manning(2019)]{Hudson2019}
Drew~A. Hudson and Christopher~D. Manning.
\newblock {GQA}: A new dataset for real-world visual reasoning and compositional question answering.
\newblock In \emph{Proceedings of the IEEE Conference on Computer Vision and Pattern Recognition (CVPR)}, 2019.

\bibitem[Hurst et~al.(2024)Hurst, Lerer, Goucher, Perelman, Ramesh, Clark, Ostrow, Welihinda, Hayes, Radford, et~al.]{hurst2024gpt}
Aaron Hurst, Adam Lerer, Adam~P Goucher, Adam Perelman, Aditya Ramesh, Aidan Clark, AJ Ostrow, Akila Welihinda, Alan Hayes, Alec Radford, et~al.
\newblock Gpt-4o system card.
\newblock \emph{arXiv preprint arXiv:2410.21276}, 2024.

\bibitem[Ji et~al.(2023)Ji, Lee, Frieske, Yu, Su, Xu, Ishii, Bang, Madotto, and Fung]{Ji2022}
Ziwei Ji, Nayeon Lee, Rita Frieske, Tiezheng Yu, Dan Su, Yan Xu, Etsuko Ishii, Yejin Bang, Andrea Madotto, and Pascale Fung.
\newblock Survey of hallucination in natural language generation.
\newblock \emph{ACM Computing Surveys}, 55\penalty0 (12):\penalty0 1--38, 2023.

\bibitem[Johnson et~al.(2017)Johnson, Hariharan, van~der Maaten, Fei-Fei, Zitnick, and Girshick]{Johnson2017}
Justin Johnson, Bharath Hariharan, Laurens van~der Maaten, Li Fei-Fei, C.~Lawrence Zitnick, and Ross Girshick.
\newblock {CLEVR}: A diagnostic dataset for compositional language and elementary visual reasoning.
\newblock In \emph{Proceedings of the IEEE Conference on Computer Vision and Pattern Recognition (CVPR)}, 2017.

\bibitem[{Kimi Team}(2025)]{kimivl2025}
{Kimi Team}.
\newblock Kimi-vl technical report.
\newblock \emph{arXiv preprint arXiv:2504.07491}, 2025.

\bibitem[Krishna et~al.(2016)Krishna, Zhu, Groth, Johnson, Hata, Kravitz, Chen, Kalantidis, Li, Shamma, et~al.]{krishna2016visual}
Ranjay Krishna, Yuke Zhu, Oliver Groth, Justin Johnson, Kenji Hata, Joshua Kravitz, Stephanie Chen, Yannis Kalantidis, Li-Jia Li, David~A Shamma, et~al.
\newblock Visual genome: Connecting language and vision using crowdsourced dense image annotations. corr.
\newblock \emph{arXiv preprint arXiv:1602.07332}, 2016.

\bibitem[Li et~al.(2023)Li, Wang, Wang, Ge, Ge, and Shan]{li2023seedbenchbenchmarkingmultimodalllms}
Bohao Li, Rui Wang, Guangzhi Wang, Yuying Ge, Yixiao Ge, and Ying Shan.
\newblock Seed-bench: Benchmarking multimodal llms with generative comprehension, 2023.

\bibitem[Li et~al.(2024)Li, Liu, Wu, et~al.]{aria2024}
Dongxu Li, Yudong Liu, Haoning Wu, et~al.
\newblock Aria: An open multimodal native mixture-of-experts model.
\newblock \emph{arXiv preprint arXiv:2410.05993}, 2024.

\bibitem[Liu et~al.(2023{\natexlab{a}})Liu, Li, Wu, and Lee]{Liu2023}
Haotian Liu, Chunyuan Li, Qingyang Wu, and Yong~Jae Lee.
\newblock Visual instruction tuning.
\newblock In \emph{Advances in Neural Information Processing Systems (NeurIPS)}, 2023{\natexlab{a}}.

\bibitem[Liu et~al.(2023{\natexlab{b}})Liu, Li, Wu, and Lee]{liu2023llava}
Haotian Liu, Chunyuan Li, Qingyang Wu, and Yong~Jae Lee.
\newblock Visual instruction tuning, 2023{\natexlab{b}}.

\bibitem[Liu et~al.(2024{\natexlab{a}})Liu, Li, Li, Li, Zhang, Shen, and Lee]{liu2024llavanext}
Haotian Liu, Chunyuan Li, Yuheng Li, Bo Li, Yuanhan Zhang, Sheng Shen, and Yong~Jae Lee.
\newblock Llava-next: Improved reasoning, ocr, and world knowledge, 2024{\natexlab{a}}.

\bibitem[Liu et~al.(2024{\natexlab{b}})Liu, Duan, Zhang, Li, Zhang, Zhao, Yuan, Wang, He, Liu, Chen, and Lin]{liu2024mmbenchmultimodalmodelallaround}
Yuan Liu, Haodong Duan, Yuanhan Zhang, Bo Li, Songyang Zhang, Wangbo Zhao, Yike Yuan, Jiaqi Wang, Conghui He, Ziwei Liu, Kai Chen, and Dahua Lin.
\newblock Mmbench: Is your multi-modal model an all-around player?, 2024{\natexlab{b}}.

\bibitem[Mostafazadeh et~al.(2016)Mostafazadeh, Misra, Devlin, Mitchell, He, and Vanderwende]{Mostafazadeh2016}
Nasar Mostafazadeh, Ishan Misra, Jacob Devlin, Margaret Mitchell, Xiaodong He, and Lucy Vanderwende.
\newblock Generating questions from images.
\newblock In \emph{Proceedings of the Annual Meeting of the Association for Computational Linguistics (ACL)}, 2016.

\bibitem[Pandey et~al.(2025)Pandey, Bodo, Phukan, and Ekbal]{pandey2025quest}
Anupam Pandey, Deepjyoti Bodo, Arpan Phukan, and Asif Ekbal.
\newblock The quest for visual understanding: A journey through the evolution of visual question answering.
\newblock \emph{arXiv preprint arXiv:2501.07109}, 2025.

\bibitem[Ren et~al.(2015)Ren, Kiros, and Zemel]{NIPS2015_831c2f88}
Mengye Ren, Ryan Kiros, and Richard Zemel.
\newblock Exploring models and data for image question answering.
\newblock In \emph{Advances in Neural Information Processing Systems}. Curran Associates, Inc., 2015.

\bibitem[Romero et~al.(2024)Romero, Lyu, Wibowo, Lynn, Hamed, Kishore, Mandal, Dragonetti, Abzaliev, Tonja, Balcha, Whitehouse, Salamea, Velasco, Adelani, Meur, Villa-Cueva, Koto, Farooqui, Belcavello, Batnasan, Vallejo, Caulfield, Ivetta, Song, Ademtew, Maina, Lovenia, Azime, Cruz, Gala, Geng, Ortiz-Barajas, Baek, Dunstan, Alemany, Nagasinghe, Benotti, D'Haro, Viridiano, Estecha-Garitagoitia, Cabrera, Rodríguez-Cantelar, Jouitteau, Mihaylov, Imam, Adilazuarda, Gochoo, Otgonbold, Etori, Niyomugisha, Silva, Chitale, Dabre, Chevi, Zhang, Diandaru, Cahyawijaya, Góngora, Jeong, Purkayastha, Kuribayashi, Clifford, Jayakumar, Torrent, Ehsan, Araujo, Kementchedjhieva, Burzo, Lim, Yong, Ignat, Nwatu, Mihalcea, Solorio, and Aji]{romero2024cvqaculturallydiversemultilingualvisual}
David Romero, Chenyang Lyu, Haryo~Akbarianto Wibowo, Teresa Lynn, Injy Hamed, Aditya~Nanda Kishore, Aishik Mandal, Alina Dragonetti, Artem Abzaliev, Atnafu~Lambebo Tonja, Bontu~Fufa Balcha, Chenxi Whitehouse, Christian Salamea, Dan~John Velasco, David~Ifeoluwa Adelani, David~Le Meur, Emilio Villa-Cueva, Fajri Koto, Fauzan Farooqui, Frederico Belcavello, Ganzorig Batnasan, Gisela Vallejo, Grainne Caulfield, Guido Ivetta, Haiyue Song, Henok~Biadglign Ademtew, Hernán Maina, Holy Lovenia, Israel~Abebe Azime, Jan Christian~Blaise Cruz, Jay Gala, Jiahui Geng, Jesus-German Ortiz-Barajas, Jinheon Baek, Jocelyn Dunstan, Laura~Alonso Alemany, Kumaranage Ravindu~Yasas Nagasinghe, Luciana Benotti, Luis~Fernando D'Haro, Marcelo Viridiano, Marcos Estecha-Garitagoitia, Maria Camila~Buitrago Cabrera, Mario Rodríguez-Cantelar, Mélanie Jouitteau, Mihail Mihaylov, Mohamed Fazli~Mohamed Imam, Muhammad~Farid Adilazuarda, Munkhjargal Gochoo, Munkh-Erdene Otgonbold, Naome Etori, Olivier Niyomugisha, Paula~Mónica Silva, Pranjal
  Chitale, Raj Dabre, Rendi Chevi, Ruochen Zhang, Ryandito Diandaru, Samuel Cahyawijaya, Santiago Góngora, Soyeong Jeong, Sukannya Purkayastha, Tatsuki Kuribayashi, Teresa Clifford, Thanmay Jayakumar, Tiago~Timponi Torrent, Toqeer Ehsan, Vladimir Araujo, Yova Kementchedjhieva, Zara Burzo, Zheng~Wei Lim, Zheng~Xin Yong, Oana Ignat, Joan Nwatu, Rada Mihalcea, Thamar Solorio, and Alham~Fikri Aji.
\newblock Cvqa: Culturally-diverse multilingual visual question answering benchmark, 2024.

\bibitem[Song et~al.(2024)Song, Yang, Li, and Shang]{song2024robust}
Lingyun Song, Chengkun Yang, Xuanyu Li, and Xuequn Shang.
\newblock A robust dual-debiasing vqa model based on counterfactual causal effect.
\newblock In \emph{Findings of the Association for Computational Linguistics: EMNLP 2024}, pages 4242--4252, 2024.

\bibitem[Srinivasan et~al.(2021)Srinivasan, Garg, Boyapaty, Tomkins, Ghadiyaram, and Pavetic]{Srinivasan2021}
Vignesh Srinivasan, Sudeep Garg, Chaitanya Boyapaty, Andrew Tomkins, Deepti Ghadiyaram, and Filip Pavetic.
\newblock {WIT}: Wikipedia-based image text dataset for multimodal pretraining.
\newblock In \emph{Proceedings of the IEEE/CVF International Conference on Computer Vision (ICCV)}, 2021.

\bibitem[Su et~al.(2018)Su, Zhu, Dong, Cai, Chen, and Li]{su2018learning}
Zhou Su, Chen Zhu, Yinpeng Dong, Dongqi Cai, Yurong Chen, and Jianguo Li.
\newblock Learning visual knowledge memory networks for visual question answering.
\newblock In \emph{Proceedings of the IEEE conference on computer vision and pattern recognition}, pages 7736--7745, 2018.

\bibitem[Team et~al.(2023)Team, Anil, Borgeaud, Alayrac, Yu, Soricut, Schalkwyk, Dai, Hauth, Millican, et~al.]{team2023gemini}
Gemini Team, Rohan Anil, Sebastian Borgeaud, Jean-Baptiste Alayrac, Jiahui Yu, Radu Soricut, Johan Schalkwyk, Andrew~M Dai, Anja Hauth, Katie Millican, et~al.
\newblock Gemini: a family of highly capable multimodal models.
\newblock \emph{arXiv preprint arXiv:2312.11805}, 2023.

\bibitem[Team et~al.(2025)Team, Kamath, Ferret, Pathak, Vieillard, Merhej, Perrin, Matejovicova, Ramé, Rivière, Rouillard, Mesnard, Cideron, bastien Grill, Ramos, Yvinec, Casbon, Pot, Penchev, Liu, Visin, Kenealy, Beyer, Zhai, Tsitsulin, Busa-Fekete, Feng, Sachdeva, Coleman, Gao, Mustafa, Barr, Parisotto, Tian, Eyal, Cherry, Peter, Sinopalnikov, Bhupatiraju, Agarwal, Kazemi, Malkin, Kumar, Vilar, Brusilovsky, Luo, Steiner, Friesen, Sharma, Sharma, Gilady, Goedeckemeyer, Saade, Feng, Kolesnikov, Bendebury, Abdagic, Vadi, György, Pinto, Das, Bapna, Miech, Yang, Paterson, Shenoy, Chakrabarti, Piot, Wu, Shahriari, Petrini, Chen, Lan, Choquette-Choo, Carey, Brick, Deutsch, Eisenbud, Cattle, Cheng, Paparas, Sreepathihalli, Reid, Tran, Zelle, Noland, Huizenga, Kharitonov, Liu, Amirkhanyan, Cameron, Hashemi, Klimczak-Plucińska, Singh, Mehta, Lehri, Hazimeh, Ballantyne, Szpektor, Nardini, Pouget-Abadie, Chan, Stanton, Wieting, Lai, Orbay, Fernandez, Newlan, yeong Ji, Singh, Black, Yu, Hui, Vodrahalli, Greff, Qiu,
  Valentine, Coelho, Ritter, Hoffman, Watson, Chaturvedi, Moynihan, Ma, Babar, Noy, Byrd, Roy, Momchev, Chauhan, Sachdeva, Bunyan, Botarda, Caron, Rubenstein, Culliton, Schmid, Sessa, Xu, Stanczyk, Tafti, Shivanna, Wu, Pan, Rokni, Willoughby, Vallu, Mullins, Jerome, Smoot, Girgin, Iqbal, Reddy, Sheth, Põder, Bhatnagar, Panyam, Eiger, Zhang, Liu, Yacovone, Liechty, Kalra, Evci, Misra, Roseberry, Feinberg, Kolesnikov, Han, Kwon, Chen, Chow, Zhu, Wei, Egyed, Cotruta, Giang, Kirk, Rao, Black, Babar, Lo, Moreira, Martins, Sanseviero, Gonzalez, Gleicher, Warkentin, Mirrokni, Senter, Collins, Barral, Ghahramani, Hadsell, Matias, Sculley, Petrov, Fiedel, Shazeer, Vinyals, Dean, Hassabis, Kavukcuoglu, Farabet, Buchatskaya, Alayrac, Anil, Dmitry, Lepikhin, Borgeaud, Bachem, Joulin, Andreev, Hardin, Dadashi, and Hussenot]{gemmateam2025gemma3technicalreport}
Gemma Team, Aishwarya Kamath, Johan Ferret, Shreya Pathak, Nino Vieillard, Ramona Merhej, Sarah Perrin, Tatiana Matejovicova, Alexandre Ramé, Morgane Rivière, Louis Rouillard, Thomas Mesnard, Geoffrey Cideron, Jean bastien Grill, Sabela Ramos, Edouard Yvinec, Michelle Casbon, Etienne Pot, Ivo Penchev, Gaël Liu, Francesco Visin, Kathleen Kenealy, Lucas Beyer, Xiaohai Zhai, Anton Tsitsulin, Robert Busa-Fekete, Alex Feng, Noveen Sachdeva, Benjamin Coleman, Yi Gao, Basil Mustafa, Iain Barr, Emilio Parisotto, David Tian, Matan Eyal, Colin Cherry, Jan-Thorsten Peter, Danila Sinopalnikov, Surya Bhupatiraju, Rishabh Agarwal, Mehran Kazemi, Dan Malkin, Ravin Kumar, David Vilar, Idan Brusilovsky, Jiaming Luo, Andreas Steiner, Abe Friesen, Abhanshu Sharma, Abheesht Sharma, Adi~Mayrav Gilady, Adrian Goedeckemeyer, Alaa Saade, Alex Feng, Alexander Kolesnikov, Alexei Bendebury, Alvin Abdagic, Amit Vadi, András György, André~Susano Pinto, Anil Das, Ankur Bapna, Antoine Miech, Antoine Yang, Antonia Paterson, Ashish
  Shenoy, Ayan Chakrabarti, Bilal Piot, Bo Wu, Bobak Shahriari, Bryce Petrini, Charlie Chen, Charline~Le Lan, Christopher~A. Choquette-Choo, CJ Carey, Cormac Brick, Daniel Deutsch, Danielle Eisenbud, Dee Cattle, Derek Cheng, Dimitris Paparas, Divyashree~Shivakumar Sreepathihalli, Doug Reid, Dustin Tran, Dustin Zelle, Eric Noland, Erwin Huizenga, Eugene Kharitonov, Frederick Liu, Gagik Amirkhanyan, Glenn Cameron, Hadi Hashemi, Hanna Klimczak-Plucińska, Harman Singh, Harsh Mehta, Harshal~Tushar Lehri, Hussein Hazimeh, Ian Ballantyne, Idan Szpektor, Ivan Nardini, Jean Pouget-Abadie, Jetha Chan, Joe Stanton, John Wieting, Jonathan Lai, Jordi Orbay, Joseph Fernandez, Josh Newlan, Ju yeong Ji, Jyotinder Singh, Kat Black, Kathy Yu, Kevin Hui, Kiran Vodrahalli, Klaus Greff, Linhai Qiu, Marcella Valentine, Marina Coelho, Marvin Ritter, Matt Hoffman, Matthew Watson, Mayank Chaturvedi, Michael Moynihan, Min Ma, Nabila Babar, Natasha Noy, Nathan Byrd, Nick Roy, Nikola Momchev, Nilay Chauhan, Noveen Sachdeva, Oskar
  Bunyan, Pankil Botarda, Paul Caron, Paul~Kishan Rubenstein, Phil Culliton, Philipp Schmid, Pier~Giuseppe Sessa, Pingmei Xu, Piotr Stanczyk, Pouya Tafti, Rakesh Shivanna, Renjie Wu, Renke Pan, Reza Rokni, Rob Willoughby, Rohith Vallu, Ryan Mullins, Sammy Jerome, Sara Smoot, Sertan Girgin, Shariq Iqbal, Shashir Reddy, Shruti Sheth, Siim Põder, Sijal Bhatnagar, Sindhu~Raghuram Panyam, Sivan Eiger, Susan Zhang, Tianqi Liu, Trevor Yacovone, Tyler Liechty, Uday Kalra, Utku Evci, Vedant Misra, Vincent Roseberry, Vlad Feinberg, Vlad Kolesnikov, Woohyun Han, Woosuk Kwon, Xi Chen, Yinlam Chow, Yuvein Zhu, Zichuan Wei, Zoltan Egyed, Victor Cotruta, Minh Giang, Phoebe Kirk, Anand Rao, Kat Black, Nabila Babar, Jessica Lo, Erica Moreira, Luiz~Gustavo Martins, Omar Sanseviero, Lucas Gonzalez, Zach Gleicher, Tris Warkentin, Vahab Mirrokni, Evan Senter, Eli Collins, Joelle Barral, Zoubin Ghahramani, Raia Hadsell, Yossi Matias, D. Sculley, Slav Petrov, Noah Fiedel, Noam Shazeer, Oriol Vinyals, Jeff Dean, Demis Hassabis,
  Koray Kavukcuoglu, Clement Farabet, Elena Buchatskaya, Jean-Baptiste Alayrac, Rohan Anil, Dmitry, Lepikhin, Sebastian Borgeaud, Olivier Bachem, Armand Joulin, Alek Andreev, Cassidy Hardin, Robert Dadashi, and Léonard Hussenot.
\newblock Gemma 3 technical report, 2025.

\bibitem[Wu et~al.(2017)Wu, Teney, Wang, Shen, Dick, and Van Den~Hengel]{wu2017visual}
Qi Wu, Damien Teney, Peng Wang, Chunhua Shen, Anthony Dick, and Anton Van Den~Hengel.
\newblock Visual question answering: A survey of methods and datasets.
\newblock \emph{Computer Vision and Image Understanding}, 163:\penalty0 21--40, 2017.

\bibitem[Yin et~al.()Yin, Fu, Zhao, Li, Sun, Xu, and Chen]{yin2306survey}
S Yin, C Fu, S Zhao, K Li, X Sun, T Xu, and E Chen.
\newblock A survey on multimodal large language models. arxiv 2023.
\newblock \emph{arXiv preprint arXiv:2306.13549}.

\bibitem[Yin et~al.(2023)Yin, Fu, Zhao, Xu, Chen, and Wang]{Yin2023}
Shukang Yin, Chaoyou Fu, Sirui Zhao, Tong Xu, Enhong Chen, and Gaoang Wang.
\newblock Woodpecker: Hallucination correction for multimodal large language models.
\newblock In \emph{Proceedings of the IEEE/CVF International Conference on Computer Vision (ICCV)}, 2023.

\bibitem[Zheng et~al.(2023{\natexlab{a}})Zheng, Chiang, Sheng, Zhuang, Wu, Zhuang, Lin, Li, Li, Xing, Zhang, Gonzalez, and Stoica]{Zheng2023_llmjudge}
Lianmin Zheng, Wei-Lin Chiang, Ying Sheng, Siyuan Zhuang, Zhanghao Wu, Yonghao Zhuang, Zi Lin, Zhuohan Li, Dacheng Li, Eric~P. Xing, Hao Zhang, Joseph~E. Gonzalez, and Ion Stoica.
\newblock Judging {LLM-as-a-judge} with {MT-Bench} and chatbot arena.
\newblock In \emph{Advances in Neural Information Processing Systems (NeurIPS)}, 2023{\natexlab{a}}.

\bibitem[Zheng et~al.(2023{\natexlab{b}})Zheng, Chiang, et~al.]{vicuna2023}
Lianmin Zheng, Wei-Lin Chiang, et~al.
\newblock {Vicuna}: An open-source chatbot impressing gpt-4 with 90\\
\newblock Tech.\ report, LMSYS, 2023{\natexlab{b}}.

\end{thebibliography}
}

\end{document}